\documentclass[10pt,twocolumn,letterpaper]{article}

\usepackage{iccv}
\usepackage{times}
\usepackage{epsfig}
\usepackage{graphicx}
\usepackage{amsmath}
\usepackage{amssymb}
\usepackage{algorithm,algpseudocode}
\usepackage{verbatim}
\usepackage{booktabs}
\usepackage{multirow}
\usepackage[pagebackref=true,breaklinks=true,letterpaper=true,colorlinks,bookmarks=false]{hyperref}

\newcommand\blfootnote[1]{%
  \begingroup
  \renewcommand\thefootnote{}\footnote{#1}%
  \addtocounter{footnote}{-1}%
  \endgroup
}

 \iccvfinalcopy % *** Uncomment this line for the final submission

\def\iccvPaperID{3552} % *** Enter the ICCV Paper ID here

\newenvironment{packed_enum}{
\begin{enumerate}
  \setlength{\itemsep}{1pt}
  \setlength{\parskip}{0pt}
  \setlength{\parsep}{0pt}
}{\end{enumerate}}

\ificcvfinal\fi
\begin{document}

%%%%%%%%% TITLE
\title{Few-Shot Generalization for Single-Image 3D Reconstruction via Priors}

\author{Bram Wallace\\
Cornell University\\
{\tt\small bw462@cornell.edu}
% For a paper whose authors are all at the same institution,
% omit the following lines up until the closing ``}''.
% Additional authors and addresses can be added with ``\and'',
% just like the second author.
% To save space, use either the email address or home page, not both
\and
Bharath Hariharan\\
Cornell University\\
{\tt\small bharathh@cs.cornell.edu}
}

\maketitle
%\thispagestyle{empty}

%%%%%%%%% ABSTRACT
\begin{abstract}
  Recent work on single-view 3D reconstruction shows impressive results, but has been restricted to a few fixed categories where extensive training data is available.
The problem of generalizing these models to new classes with limited training data is largely open.
To address this problem, we present a new model architecture that reframes single-view 3D reconstruction as learnt, category agnostic refinement of a provided, category-specific prior.
The provided prior shape for a novel class can be obtained from as few as one 3D shape from this class.
Our model can start reconstructing objects from the novel class using this prior without seeing any training image for this class and without any retraining.
% This model yields a 6\% boost in novel category performance over a category-agnostic baseline using only 1-shot knowledge of novel category shapes with no images required and no retraining and almost doubles this improvement with further information.
Our model outperforms category-agnostic baselines and remains competitive with more sophisticated baselines that finetune on the novel categories.
Additionally, our network is capable of improving the reconstruction given multiple views despite not being trained on task of multi-view reconstruction.
\end{abstract}

%%%%%%%%% BODY TEXT
\section{Introduction}
\blfootnote{© 2019 IEEE.  Personal use of this material is permitted.  Permission from IEEE must be obtained for all other uses, in any current or future media, including reprinting/republishing this material for advertising or promotional purposes, creating new collective works, for resale or redistribution to servers or lists, or reuse of any copyrighted component of this work in other works.}

A key aspect of visual understanding is recovering the 3D structure of a scene. 
While classically such recovery of 3D structure has used multiple views of a scene, there has been recent research on 3D reconstruction \emph{from a single image} using machine learning techniques.
However, recovering 3D structure from a single image is a challenging learning problem.
First, the output space is not just very large (e.g., represented as voxels, a $100 \times 100 \times 100$ grid is already a million-dimensional space) but also very \emph{structured}: of all possible 3D shapes that are consistent with an image of a chair, a vanishingly small number are valid chair shapes.
To perform well, the machine learning algorithm needs to capture a prior over possible chair shapes.
Large, deep networks can indeed capture such priors when provided enough chairs for training, and this has been the dominant approach taken by prior work.
However this leads to the second challenge: the cost of acquiring training data.

\begin{figure}[h]
\begin{center}
\includegraphics[width=\linewidth]{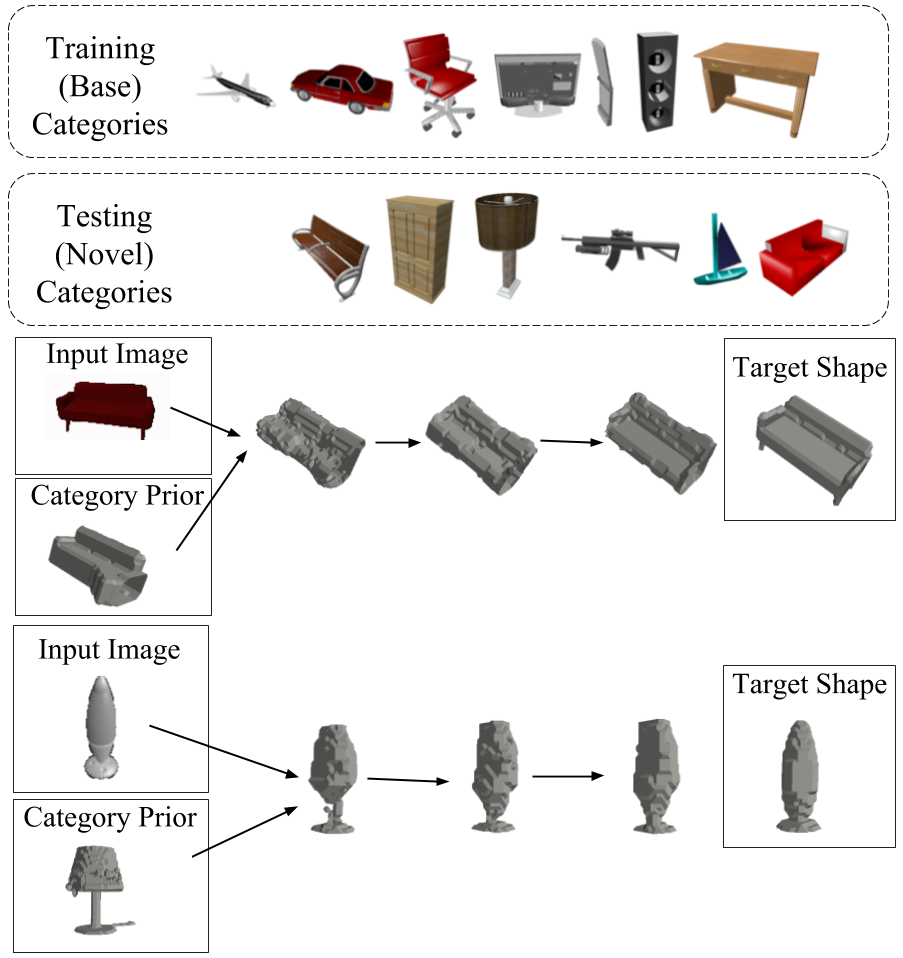}
\end{center}
\caption{We train on 7 base categories and test the model's few-shot transfer ability on 6 novel categories. Our model takes in an image of the object to reconstruct along with a category-specific prior shape which can be as simple as a single novel class example. It then optionally iteratively refines this prior to produce a reconstruction.}
\label{fig:voxel_evo}
\end{figure}

Training data for single view 3D reconstruction requires either 3D shapes~\cite{r2n2} or at the very least multiple views of the same physical object~\cite{yang}.
Such training data can be acquired for a small number of categories, but is too expensive to obtain for every single object class we might want to reconstruct. 
Prior work attempts to circumvent this issue by training a \emph{category-agnostic} model~\cite{ptn, yang}, but such models might underperform due to ignoring category-specific structure in the output space.
Therefore, we ask: is it possible to capture category-specific shape priors for single-image 3D reconstruction from very limited training data?

In this paper, we show that the answer is yes.
We present a simple \emph{few-shot transfer learning} approach that can very quickly learn to reconstruct new classes of objects using very little training data.
%
%Is it possible to learn effective  
%Reconstructing a 3D model from images is a challenging problem that has seen much development in recent years \cite{yang, r2n2, fan, pix2mesh}. The dataset of choice for this task is ShapeNet, which offers high-quality renderable 3D models \cite{shapenet}. Most work has focused on a dozen or so defined categories, with very little focus on performance on non-training categories. To our knowledge there has been almost no work specifically on the problem of few-shot learning in this context.
%
%Few-shot learning is notoriously difficult in deep learning due to the large parameter space. Striking a balance between leveraging information and overfitting is difficult and normal training techniques do not work well. The need for few-shot learning can arise due to the large expense in acquiring suitable data to train upon \cite{vinyals, snell, ravi}. 3D models very difficult to acquire and also require accompanying images.  They also are very large files taking up much storage space. Few-shot learning allows for low time and memory cost while performing transfer learning.
Instead of training a direct mapping from RGB images to 3D shapes, we train a model that uses image input to \emph{refine} an input prior shape.
This simple reparametrization allows us to swap in new priors for new classes \emph{at runtime}, enabling single view reconstruction of novel object classes \emph{with no additional training}.
We show that this boosts reconstruction accuracy significantly over category-agnostic models.

We find an additional benefit to implementing the prior in this way: the output of our model can used as a new prior and fed back into the model to iteratively refine the prediction. 
While the notion of iterative prediction for better accuracy has been proposed before~\cite{autocontext, inference-machines, iterative-error-feedback}, the connection to few-shot learning in this context is new. 
We demonstrate that this iterative strategy can also be used out-of-the-box for competitive multi-view reconstruction \emph{without any multi-view training}.
Our approach is shown in Figure~\ref{fig:voxel_evo}.

Summarizing the contributions of this paper, we:
\begin{packed_enum}
\item Propose an augmented network architecture and training protocol that can incorporate prior categorical information at runtime
\item Demonstrate this network's ability on few-shot learning
\item Demonstrate this network's ability to perform competitive multiview reconstruction without being trained on the task
\end{packed_enum}

\section{Related Work}
Classically, the problem of 3D reconstruction has been solved using multiple views and geometric or photometric constraints~\cite{pmvs, spacecarving}.
However, recently the success of convolutional networks on recognition tasks has prompted research into using machine learning for \emph{single-view} 3D reconstruction.
Early success in this paradigm was shown by R2N2~\cite{r2n2}.
R2N2 iteratively refines a 3D reconstruction based on multiple views; this is similar in spirit to our approach of refining a prior shape, but the focus is on multi-view reconstruction and not generalization.
Later work has since improved the underlying representation of 3D shapes~\cite{atlasnet, cmrKanazawa18, foldingnet, pix2mesh, tatarchenko2017octree, wang2017cnn}, replaced 3D training data with multiple calibrated views of each training object~\cite{mvcTulsiani18, ptn, yang}, incorporated insights from geometry to improve performance~\cite{kar2017learning, nips_generalization}, or made other improvements to the learning procedure~\cite{wu2017marrnet, wang20173densinet}. 
However, the question of generalizing to novel classes with limited training data is under-explored.

Work on generalization in the context of 3D reconstruction is limited.
Recently Tatarchenko et al. demonstrated that single view 3D reconstruction models tend to memorize and retrieve similar shapes from the training set; an indication of overfitting~\cite{tatarchenko2019single}.
This suggests that more generalizable models are necessary.
Yang et al. are one of the first to attempt transfer learning for 3D reconstruction and find the best solutions to be using class-agnostic models and finetuning.~\cite{yang}. 
% Finetuning requires offline time and results in a model tailored to solely the new category.
We show that our approach outperforms both of these solutions when training data for novel classes is limited.
Class agnostic models might be more generalizable if they incorporate geometrical constraints~\cite{nips_generalization} or leverage pose information\cite{lsm}. 
This idea of using geometry is orthogonal to, and indeed complementary to, our insight of separating out the category-specific prior.

The notion of using or learning priors has also been explored before.
One approach to using priors is to use an adversary to enforce realistic reconstruction \cite{wu_shape_priors, view_priors}.
Cherabier et al. use shape priors to learn from relatively little data, but focus on scene reconstruction with semantically labeled depth maps as inputs \cite{cherabier}. 
3D-VAE-GAN is similar to our work in leveraging categorical knowledge\cite{vaegan}. 
Closer in spirit to our work are single-view reconstruction methods that use meshes as their underlying representation, which often function by deforming a prior mesh~\cite{pix2mesh,cmrKanazawa18}
However, in all these approaches, the focus is on improving the in-category performance rather than on generalization or transfer; which are often not even evaluated.
In contemporary work, Wang et al. propose to deform a source mesh to match a taarget shape, but their focus is on point cloud registration rather thaan single view reconstruction~\cite{wang20193dn}.

The approach we propose also has connections to models which use iterated inference in structured prediction problems.
This idea was originally proposed for more classical approaches based on graphical models~\cite{autocontext, inference-machines} but has recently been applied to deep networks~\cite{iterative-error-feedback}.
An iterative approach to single-view reconstruction is that of Zou et al., who build reconstructions via sequential concatenation of shape primitives \cite{shape_primitives}.
Although shape primitives can sometimes lack expressivity for complex shapes, they also capture some priors about shape.

%Pix2Mesh and Fan et al. both work on the problem of single-view 3D reconstruction, but neither touch on generalization. 

Our work is also related to few-shot transfer learning.
Most prior work on the few-shot learning problem focuses on classification tasks.
A large class of recent work in this domain uses the idea of meta-learning, where the model is trained using simulated few-shot learning scenarios it will encounter in deployment~\cite{vinyals, SnellNIPS2017, WangCVPR2018}.
Our training procedure is similar in this regard, but focuses on structured prediction instead of classification.
Some early work on few-shot learning also has the notion of separating out a class-specific mean shape from class-agnostic aspects of the classification problem~\cite{MillerCVPR2000}, but again the key difference in this paper is the structured prediction problem.

\section{Problem Setup}
We are interested in learning single view 3D reconstruction for novel classes from very limited training data.
We approach this broad problem through the lens of \emph{few-shot learning} and \emph{transfer learning}.
We assume that we have a large dataset of 3D shapes with corresponding images for some set of classes, which we call \emph{base classes}~\cite{Hariharan_2017_ICCV}.
We will train our model on these base classes.

After training, the model will encounter \emph{novel classes} for which we have very limited ground truth 3D shape information.
In general, we will assume access to between 1 and 25 examples of 3D models for each class.
Note that we do not assume that these 3D models come equipped with corresponding images; the model we propose below only uses the 3D models themselves to construct a category-specific prior.
The model must use these example 3D models to reconstruct 3D shape for \emph{test examples} of each class.
The final performance metric will be its reconstruction accuracy on these test examples. 
In particular, we follow prior work and look at intersection-over-union between the predicted shape and the ground truth.

\begin{figure*}[th]
\begin{center}
\includegraphics[width=0.8\linewidth]{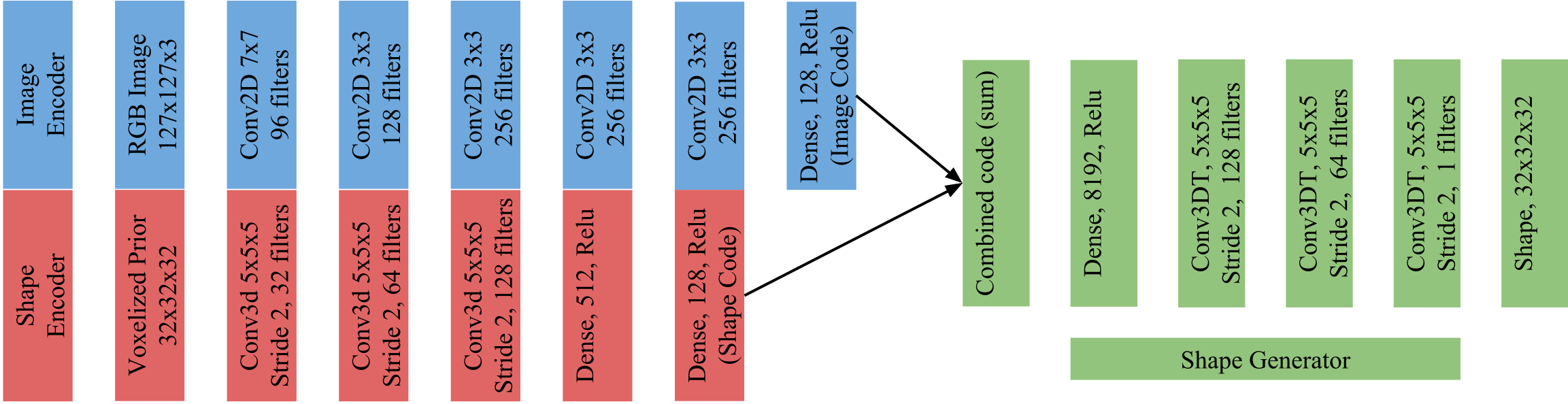}
\end{center}
   \caption{Our model is dual-input. The first input is an image encoded using the exact same architecture as 3D-R2N2 \cite{r2n2}. The second is a voxelized prior shape encoded via 3D convolutions, similar to Yang et al. \cite{yang}. The generator is similar to that of Yang et al. The 128-dimensional output of the encoders are summed. Each Conv2D layer is followed by 2x2 MaxPooling and LeakyRelu with $\alpha=0.01$ and each Conv3D layer is followed by LeakyRelu with $\alpha=0.3$. ReLu activations are used in the generator.}
\label{fig:network_diagram}
\end{figure*}

\section{Approach}
\subsection{Model Architecture}\label{sec:model}
We first create a \emph{category-specific shape prior} in the form of a voxel grid by averaging the voxel representations for a small number of 3D shapes available for the novel class.
%We now describe our proposed solution to this problem.
Note that the individual voxels can take floating point values in this grid.
We then design a \emph{category-agnostic} neural network that \emph{refines} this category-specific prior based on image input.
%This neural network takes two inputs: an image and the prior.
%It uses two encoders to encode the image and the category prior into a common embedding space.
This neural network uses two encoders to encode the image and the category prior into a common embedding space.
The embeddings of the image and the category prior are added together and fed into a decoder that produces the refined shape.

This scheme for few-shot prediction offers several major advantages:
\begin{packed_enum}
\item Very little runtime is required to incorporate the few-shot information. The shapes must simply be loaded and averaged, a negligible operation compared to the network's forward pass.
\item No retraining of the network is performed. % , all parameters stay the same.
\item There is no difference in the predictive method for new or old categories. 
\item Multiple types of priors can be incorporated in this fashion.
\item No corresponding images are required for transfer learning, only shapes. These might be obtained from CAD models created by designers. 
\end{packed_enum}

\noindent\textbf{Iterative prediction:} 
Because our model refines an input shape, its output can be fed back in again to refine the shape further. 
Such iterative refinement has been shown to be useful for structured prediction problems~\cite{autocontext, inference-machines, iterative-error-feedback}. 
We evaluate both iterative and non-iterative versions in our experiments.

\noindent\textbf{Implementation details:}
The precise architecture is shown in Figure~\ref{fig:network_diagram}. The image encoder takes in a $127 \times 127$ RGB image as input and feeds it through a series of convolutions ($3 \times 3$ except for an initial $7 \times 7$) alternating with max pooling layers and finishing with a fully connected layer. The shape encoder takes in the category prior as input. The shape encoder is a series of 3-dimensional convolutions followed by two dense layers. The image encoder is the same as that used by R2N2 and the shape encoder and decoder are similar to architectures employed by Yang et al~\cite{yang}. The output of the two encoders are feature vectors of length 128 which are summed before being fed into the generator.

 LeakyRelu is used in both encoders, with $\alpha=0.01$ in the image encoder and $\alpha=0.3$ for the shape half \cite{activations}. Traditional Relu was used in the shape generator.  A sigmoid activation is applied at the final step of the shape generator. % to squash the values into the $(0,1)$ range.

\begin{figure}[h]
\begin{center}
   \includegraphics[width=\linewidth]{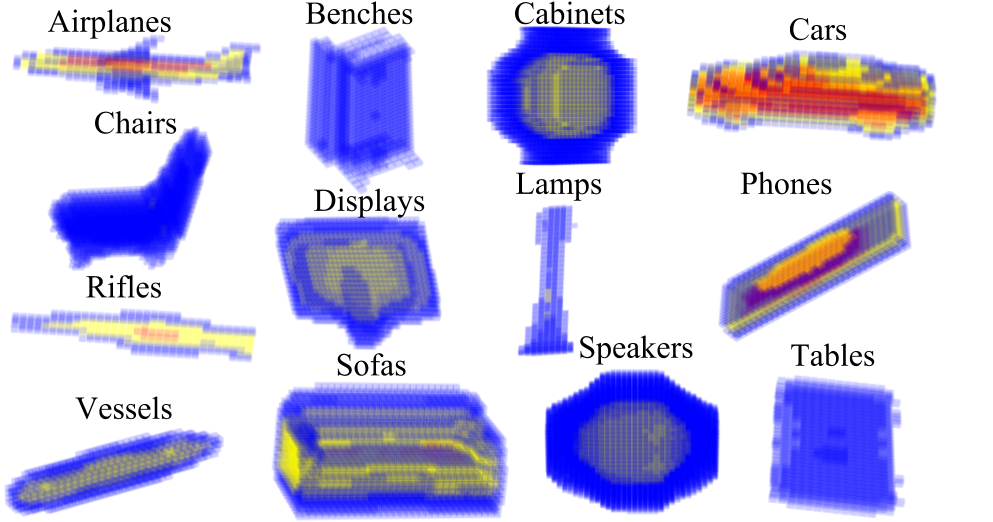}
\end{center}
   \caption{The averaged shapes of the entire training dataset for each category. The color represents the frequency of models in which a given gridpoint was occupied. Red indicates 90-100\%, yellow 60-90\% and blue 30-60\%. We see that  \textbf{airplanes},  \textbf{cars}, and  \textbf{rifles} have an extremely consistent shape, while other categories such as  \textbf{lamps} and  \textbf{tables} have relatively weak priors, with no visible non-blue gridpoints.}
\label{fig:average_voxels}
\end{figure}

\subsection{Training}\label{sec:training}
% We train the model by simulating during training what it will observe during testing. 
For every training datapoint, we sample an image from one of the base classes and the corresponding ground truth 3D shape as the target.
Our secondary input, the prior shape, consists of an average of some number of other same-category shapes from the training set. For some models, this prior shape is the ``Full Prior": all the shapes in the train dataset are averaged. When a ``k-shot" prior is used, it consists of $k$ averaged shapes, always from the training dataset. The ``Full Prior" models always have the same initial input shape within a category while the ``k-shot" prior networks use a different randomly generated prior for each image-target pair. We display the ``Full Prior" shapes for each category in Figure~\ref{fig:average_voxels}.
The loss is the binary cross-entropy loss.

\noindent\textbf{Training iterative models:}
To train the model in an iterative setting, we repeat each training batch multiple times, with the model outputs of one iteration being fed as inputs in the next.
For each batch from the generator, the same input images and target shapes are used each time and the input shapes change after the first step, being the output of the previous forward pass (Algorithm~\ref{algo:batch_iteration}).

\noindent\textbf{Implementation details:}
All experiments are done using Keras with a Tensorflow backend \cite{keras, tensorflow}. Training is done in batches of 32 using an Adadelta optimizer \cite{adadelta}. 
Early stopping is used, with our metric of accuracy being the per-category average Intersection-over-Union (IoU) on the base classes with a threshold on the output of $0.4$. 
This threshold is standard in literature, and in our case as well offered good performance.
 Relative performances of the models were maintained at different thresholds.

\begin{comment}
\begin{figure}[h]
\begin{center}
   \includegraphics[width=0.8\linewidth]{figures/shapenet_data}
\end{center}
   \caption{Examples from the Shapenet dataset. 127x127 RGB images on the left and corresponding 32x32x32 voxelized shapes on the right. Shapes are rendered as meshes using the marching cubes algorithm.}
\label{fig:shapenet_data}
\end{figure}
\end{comment}

\begin{algorithm}
  \begin{algorithmic}[1]
  \For{epoch in epochs}
   \For{batch in batches}
     \State Load input images, input shapes, target shapes from generator
     \For{$iter\_i$ in $1..\#iters$}
          \State Train on  input images, input shapes, target shapes with backprop \label{algo:train_op}
          \State Set the input shapes equal to the output of the model
     \EndFor
   \EndFor
\EndFor
  \end{algorithmic}
\caption{Training for iterative refinement.}
  \label{algo:batch_iteration}
\end{algorithm}

\section{Results}\label{section:results_intro}
\subsection{Experimental setup}
We experiment with the ShapeNet dataset. 
Seven of the 13 categories are designated \emph{base classes} and are used during training: \textbf{airplanes, cars, chairs, displays, phones, speakers,} and \textbf{tables} (matching the work of Yang et al~\cite{yang}). 
We use $127 \times 127$ RGB rendered images of models and $32 \times 32 \times 32$ voxelized representations. Examples of the data as input-target pairs can be seen in Figures~\ref{fig:voxel_evo} and \ref{fig:example_predictions}. 
%In total we use approximately 45k models from 13 different categories. 
Each model has 24 associated images from random viewpoints. 
We use the same training-testing split as R2N2 which was an 80-20 split. We further divide this into a 75-5-20 split to obtain a validation set.

%There are two parts of the  testing process that must be explained: the use of priors and the iterative scheme. 
%First we discuss the prior. 
When testing on base classes, we use the full prior unless otherwise noted. 
For novel-category testing, we always report the number of shapes being averaged into the prior. %, usually this number is between 1 and 25. 
We consider both iterative and non-iterative models.
%Iterative evalution can be performed using Algorithm~\ref{algo:batch_iteration} with Operation~\ref{algo:train_op} omitted. 

\paragraph{Baselines:}
We compare against multiple baselines. 
The first baseline is a category-agnostic mapping from images to 3D shapes.
This model uses the same image encoder and shape decoder architecture, but does not use any category-specific prior as input or employ novel-category data at all.
Such a category-agnostic model has been shown to perform very well in prior work~\cite{r2n2, yang} and is thus a strong baseline.
The second baseline \emph{finetunes} the image-only model on the novel classes.
$k$ shapes are rendered from up to 24 viewpoints, resulting in between $k$ and $24k$ image-pairs (depending on the model) which are then finetuned on.
% We experiment with finetuning either the entire model or solely the generator on the provided novel-category shapes \emph{each paired with an associated image}.
Note that this baseline uses paired images, which are not available to our approach.
We finetune the models for 200 iterations using SGD with a learning rate of 0.005.

\subsection{Main results}

\begin{table}[t]
\begin{center}
\resizebox{0.99\linewidth}{!}{
\begin{tabular}{ c l  c c c c}
\toprule
\# Novel class          & Image-Only     &   Finetune         & Finetune               & Finetune               & 1-Iteration      \\
Examples ($k$)                 & Baseline         &  1 Render    & 5 Renders  & 24 Renders                     & 1-Shot         \\
% \cmidrule(l{15pt}r{40pt}){6-8} \cmidrule(l{15pt}r{25pt}){9-10}
\midrule
1-shot                                &   0.36      & {0.38}    & {0.38}                      & 0.39                         &    {0.38}                       \\
2-shot                                 & 0.36  &  0.38                   & 0.39                         & 0.40                     &    {0.39}      \\    
3-shot                                 & 0.36   &  0.38                  & 0.39                         & 0.41            &    {0.39}                   \\
4-shot                                &   0.36    &  0.39           & {0.40}                & 0.42                &  {0.39}                         \\
5-shot                                &   0.36    &  0.39                   & 0.40                      &0.42      &      {0.40}                           \\
10-shot                                 & 0.36   &  0.39                  & 0.42                        & 0.44             &    {0.40}                    \\
% 25-shot                                 &  0.36   & 0.41                & \textbf{0.41}                                        &     0.40                                              \\
Full Prior                               &  0.36   &                            &                                                         &  0.40                                                 \\
\bottomrule
\end{tabular}
}
\end{center}
   \caption{Few-shot learning results on novel classes. The Image-Only Baseline does not incorporate new-category information at all.
% For the "Full Prior" row in the finetuning columns the entire category's training dataset was fitted to. This serves as an upper-bound on performance. Both of these are \emph{fully supervised} approaches.
The ``1-Iteration 1-shot" model is a non-iterative model trained with 1-shot priors and tested with priors consisting of $k$ averaged shapes from the training category.
% On each row, the result of the top model(s) is \textbf{bolded}.
We see that our model offers competitive performance, especially in very low-shot regimes, despite \emph{no image supervision or retraining}.
Scores reported are category-wise average IoU.
The same Image-Only Baseline architecture achieved 0.55 IoU when trained on \emph{all} of the classes at once.
We perform 3-5 runs of each experiment with $\sigma_{IoU}<0.01$.}
\label{table:few_shot_summary}
\end{table}

\begin{figure}[b]
\begin{center}
\includegraphics[width=0.8\linewidth]{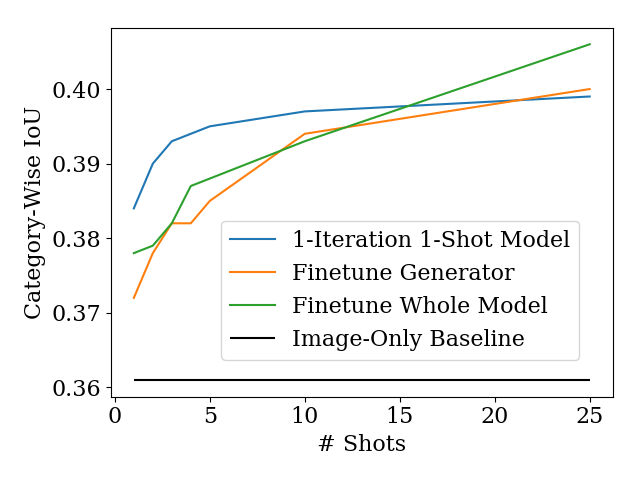}
\end{center}
\caption{Performance of the 1-iteration 1-shot-trained model against various baselines tuned with 1 view per model. We see that the majority of improvement (60\%) comes within the first 1 to 3 shots.}
\label{fig:num_shot_vs_ious}
\end{figure}

%We first present results for multiple model variants under the few-shot learning setting in Table~\ref{table:few_shot_summary}. 
We first present results for our best model variant under the few-shot learning setting along with multiple baselines in Table~\ref{table:few_shot_summary}. 
We vary the number of novel-class example shapes available and evaluate models on the average IoU across all novel classes.
Figure~\ref{fig:num_shot_vs_ious} plots the performances of the models for priors containing varying amounts of information. 

\begin{figure}[t]
\begin{center}
   \includegraphics[width=0.8\linewidth]{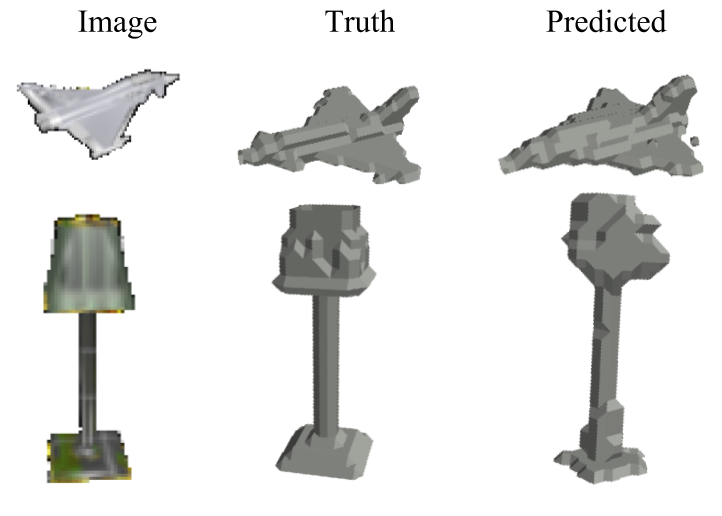}
\end{center}
   \caption{Examples of ground-truth and predicted shapes from an image. Note that the category \textbf{lamps} is not in our training set, we use a prior to enable generalization to this previously unseen category.}
\label{fig:example_predictions}
\end{figure}

We observe that our best model variant (1 iteration trained on 1-shot priors) significantly outperforms the category-agnostic baseline across the board on the novel classes indicating the usefulness of the category prior.
Compared to the finetuning-based approaches, our method outperforms the variant that sees one rendering per model, and is competitive with the variant that sees five images per model.
Note that the finetuning approaches see significantly more information than our approach, which gets \emph{no novel-class images at all}.
Furthermore, unlike the finetuning approaches, our model \emph{requires no retraining on the target class at all}.
Any new class can thus be added to our models' repertoire simply by adding a corresponding prior.
% We also see that our model outperforms both finetuning approaches for $k \le 10$ \emph{despite not retraining the model and having no access to novel images}.
%We also see that our model remains competitive with finetuning approaches despite not retraining the model and having no access to novel images.
Furthermore, as shown in Figure~\ref{fig:num_shot_vs_ious}, we find that \emph{very few novel-class shapes} are needed for this prior: with only 5 shapes, our model sees a gain of 4 points over the category-agnostic baseline.
Example predictions are shown in Figure~\ref{fig:example_predictions}.

We include results from additional variants of our model in Table~\ref{table:few_shot_supp}.
We note that among the different model variants, models that perform iterated inference do not outperform the 1-iteration 1-shot model.
Furthermore, for more informative priors, iteration buys no gain and sometimes even hurts the novel classes.
Despite these shortcomings, we do find that they prove useful in the multi-view setting (Sec.~\ref{sec:multiview}).

%Iteration buys no gain for the base classes, and sometimes hurts the novel classes.
%Despite these shortcomings, we do find that they prove useful in the multi-view setting

\begin{table}[t]
\begin{center}
\resizebox{0.99\linewidth}{!}{
\begin{tabular}{ c l c c c c}
\toprule
\# Novel class        & 1-Iteration     &   2-Iteration                                    & 3-Iteration                                                                &     2-Iteration      \\
Examples                        & Full Prior                 & Full Prior                                      & Full Prior                                                                     &    1-Shot Prior\\
\midrule
1-shot                                   &     0.34                      &      0.36/\underline{0.37}                     &      0.34/0.37/{\underline{0.38}}                     &  \underline{{0.38/0.38}}      \\
2-shot                                     &   0.36                      &   \underline{0.38/ 0.38}                        &       0.37/{\underline{0.39}}/0.38                     &  \underline{{0.39}}/0.38     \\    
3-shot                                     &  0.36                         &  \underline{0.38/0.38}                 &     0.38/\underline{{0.39/0.39}}                        &   \underline{{0.39}}/0.38   \\
4-shot                                     &   0.36                         &    \underline{{0.39}}/0.38     &   \underline{{0.39/{0.39}/{0.39}}}           &   \underline{{0.39}}/0.38                \\
5-shot                                    &   0.37                         &   \underline{0.39}/{{0.38}}               &  \underline{0.39/0.39/0.39}                                    &  \underline{0.39}/0.38                  \\
10-shot                                  &  0.37                         & \underline{0.39/0.39}                      &  \underline{{0.40/0.40}}/0.39                             &   \underline{0.39}/0.38                 \\
25-shot                                   &  0.37                         &  \underline{0.40}/0.39                       &     \underline{0.40/0.40}/0.39                                    &   \underline{0.40}/0.38                 \\
Full Prior                                  &   0.37                         &  \underline{0.40}/0.39                     &    {\underline{0.41}}/0.40/0.39                      &    \underline{0.40}/0.38                 \\
\bottomrule
\end{tabular}
}
\end{center}
   \caption{Few-shot learning results on novel classes for additional model variants. Models are trained and tested for the same number of iterations. Setup as in Table~\ref{table:few_shot_summary}. The best-performing iteration for each model is \underline{underlined}.}
\label{table:few_shot_supp}
\end{table}

 %In Table~\ref{table:training_summary} we see that our models do indeed have comparable peformance to the baseline.
In Table~\ref{table:training_summary} we see that the improvements on novel classes do not come at the expense of performance within base classes.
We also include the averaged performance of the R2N2 network as presented in the original work, to show that our baseline when trained on all 13 categories is slightly better, and thus a very strong control architecture to use. 
% (while R2N2 is no longer the state-of-the-art, the work we present here is focused on a conceptual framework for few-shot learning and semi-supervised multi-view reconstruction that could be applied to multiple architectures).

% While this might suggest that iterative models are not useful, we find that they show benefits in the multi-view setting.
%

\begin{table}[h]
\begin{center}
\resizebox{0.9\linewidth}{!}{
\begin{tabular}{ l l l }
\toprule
Training Procedure & Training & Base-class  \\  
& Classes & performance\\
\midrule
R2N2 & All & 0.58   \\
1 Iteration No Prior & All & 0.59 \\
\midrule
1 Iteration No Prior &  Base &    0.62   \\
1 Iteration Full Prior & Base & 0.63 \\ 
2 Iteration Full Prior & Base & 0.62/0.62   \\ 
3 Iteration Full Prior & Base & 0.61/0.61/0.61    \\
1 Iteration 1-Shot Prior & Base & 0.62 \\
2 Iteration 1-Shot Prior & Base &  0.61/0.61  \\
\bottomrule
\end{tabular}
}
\end{center}
   \caption{Training Category Results Summary. Models are tested on the test dataset of the training categories. The prior used is the same as during training. Our models perform comparably to an image-only baseline fitted on the training categories. This baseline outperforms R2N2 substantially, which we see is primarily due to the reduced categorical load.}
\label{table:training_summary}
\end{table}

%We see that while the performances improve incrementally in the limit, the vast majority of the improvement comes in the first several shots. Specifically, the 1-iteration 1-shot model improves by 0.009 IoU going from a prior containing 3 vs 1 shapes, then subsequently only gains improvement of 0.006. Similarly, the 3-iteration model improves by 0.013 in this span, and then 0.008 in the remainder. These improvements are in addition to 1-shot performance gaps of 0.023 and 0.015 to the baseline respectively. This shows that our framework is particularly powerful in the 1-to-3-shot context, which is a rare attribute. 

\subsection{Multi-view Reconstruction}
\label{sec:multiview}
In practice, it is often the case that we have more than one view of the object we want to reconstruct.
Neural network approaches to multi-view reconstruction from uncalibrated views typically use recurrent neural networks as in R2N2.
However, since our model is framed as \emph{refining} a prior, we can use it iteratively, feeding in new images at each step.

Table~\ref{table:multiview} shows the performance of two of our best variants in the multi-view settings for both the base and the novel classes.
We show both the non-iterative model trained on 1-shot priors (best performer in Table~\ref{table:few_shot_summary}) as well as the 3-iteration model trained on the full prior.
For the base classes, we use the full category prior and compare to R2N2 (with the caveat that R2N2 is trained on more classes).
For the novel classes, we use a 1-shot prior.

We find that on the base classes, our 3-iteration model significantly improves on its single-view accuracy and achieves competitive performance to R2N2 \emph{without any multi-view training}.
Access to multiple views is even more beneficial for novel classes, where performance increases by close to 7 points.
%While our single-view 3-iteration model achieved a maximum IoU of 0.38 when given a 1-shot prior, we see that multiple views can boost the IoU by 0.03, an over 7\% increase. 
This again is despite not being trained on the multiview task, and only being given 1 example shape to learn from. 
%To our knowledge this is the first such system to peform competitive multiview reconstruction despite not being trained on the task.

Interestingly, the non-iterative model is unable to benefit from the additional images. This suggests that when the target task requires iterative refinement, \emph{training} for iterative refinement might be necessary, even if it is only single-view training.
\begin{table}[h]
\begin{center}
\resizebox{0.99\linewidth}{!}{
\begin{tabular}{ l  c  c  c  c  c }
\toprule
Method & \# Views=1 & 2 & 3 & 4 & 5  \\  
\midrule
\emph{Base classes} &  &  &  &  & \\
\;\; R2N2 \cite{r2n2} & 0.58 & 0.62 & 0.64 & 0.64 & 0.65  \\
\;\; LSM \cite{lsm} & 0.60 & 0.71 & 0.75 & 0.77 & -  \\
% Pix2Vox\* \cite{pix2vox} & 0.658 & 0.682 & 0.692 & 0.695 & 0.697 \\
\;\;3-Iteration & {0.61} & {0.63} & {0.63} & {0.63} &  {0.64}  \\
\;\; 1-iteration 1-shot & 0.62 & 0.62 & 0.62 & 0.62 &  0.62 \\
\midrule
\emph{Novel classes} &  &  &  &  & \\
\;\;3-Iteration & {0.34} & 0.38 & 0.40 & 0.40 &  0.41   \\
\;\;1-iteration 1-shot & 0.39 & 0.39 & 0.39 & 0.39 & 0.39 \\
\bottomrule
\end{tabular}%
}
\end{center}
   \caption{ Multi-view performance (IoU) on base categories (top) and novel categories (bottom). For base classes we compare to R2N2 (of which our architecture is an augmented version) and Learned Stereo Machines (an approach which uses provided pose information to backproject the pixels into a canonical, shared, reference frame. A full prior is used for the base classes and a 1-shot prior is used for the novel classes. The models iterative scheme can be adapted to multi-view reconstruction and shows substantial benefit despite not being trained on the task. }
   \label{table:multiview}
\end{table}

%We see that our full-prior 3-iteration model improves substantially on its training-category IoU of 0.611. This is quite remarkable, as our model is not trained using multiple images at any step, in Algorithm~\ref{algo:batch_iteration} the input images are only changed when a new batch is drawn from the generator.While our training dataset had 24 images associated to each object, they were only viewed in single image-object pairs. Theoretically our model could have been trained on a dataset with only one image associated to each 3D object and still possibly achieved the same multi-view results. To our knowledge this is the first such system to peform competitive multiview reconstruction despite not being trained on the task.

\subsection{Analysis}
As shown above, our approach demonstrates strong performance on novel classes with very limited training data, for both single view and multiview reconstruction.
We now perform a thorough analysis of our results, including the following questions:
\begin{packed_enum}
\item How do the performance improvements break down over categories and over examples?
\item How important is the prior?
\item Can this approach be used on real-world images?
% \item What are the gains, if any, of iterative refinement?
%\item Comparing known-category performance across models and to the baseline % done
%\item Measuring the performance on multi-view reconstruction % done
%\item Performance of naively guessing the input % done
%\item Graphing the performance across iterations % done
%\item Evaluating the performance of the models given an incorrect prior % done 
%\item Comparing the distributions of performance between models
%\item Evaluating per-category transfer performance
% Wasn't able to find why individual categories differed
% Possible, different # of categories?
% Some kind of per-category comparison
\end{packed_enum}
\subsubsection{Analyzing Performance Distribution}\label{section:performance_dist}

\begin{figure}[t]
\begin{center}
   \includegraphics[width=0.9\linewidth]{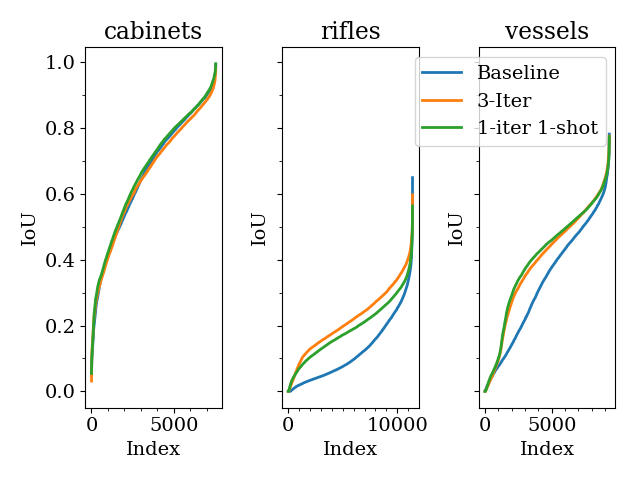}
\end{center}
   \caption{Here we plot the IoUs in increasing order for each model-category pair. We see that both of our new models substantially outperform the baseline on \textbf{rifles} and \textbf{vessels}. A 10-shot prior was used. Note that this is the same data as shown in Figure~\ref{fig:iou_scatters}.}
\label{fig:iou_cdf}
\end{figure}

While we have presented the average IoU on transfer learning tasks, this doesn't address the question of how these statistical results are achieved (e.g. translation of the distribution or a few exceptionally strong reconstructions). 
% Such a shift in mean could be caused by a general translation of the distribution, very poor reconstructions on the baseline being corrected, or some reconstructions in the new model being exceptionally strong. 
%To distinguish between these three hypotheses, we first look at the error distributions by plotting the IoUs for three categories and models in Figure~\ref{fig:iou_cdf}. Here a full prior is used.
To determine the cause, we first look at the error distributions by plotting the IoUs for three categories and models in Figure~\ref{fig:iou_cdf}. Here a 10-shot prior is used.

\begin{comment}
\begin{figure}[t]
\begin{center}
   \includegraphics[width=0.9\linewidth]{figures/category_hists.png}
\end{center}
   \caption{IoU histograms of test points for each category across different models. We see that the new models incorporating priors have less extremely poor guesses for \textbf{rifles} and \textbf{vessels} leading to increased accuracy. A full prior was used for both of our models.}
\label{fig:iou_dist}
\end{figure}
\end{comment}

We see that increases in accuracy primarily come not from substantially increasing the number of highly accurately reconstructed shapes, but reducing the number of poorly reconstructed shapes. 
In  \textbf{rifles} for example, the baseline has an IoU of less than 0.1 for over half the instances, whereas for our models this number is less than 17\%.

 %The \textbf{rifles} class is very difficult to reconstruct, with extremely low IoUs. This results in the baseline distribution having a very flat initial curve while our model curves instead have a very steep slope at the beginning. The resulting gap is maintained throughout the plot, only tapering to a close at the far-right endpoint. In the \textbf{vessels} category the baseline curve is nearly linear, representing a near-uniform distribution. The new model curves, however, again have sharp initial improvement, demonstrating very few poor reconstructions. This shows that our architecture reduces the number of critical errors, a very important trait. It is interesting to note that despite having much higher average peformance our model does not produce many extraordinarily high individual results. 

\begin{figure}[h]
\begin{center}
   \includegraphics[width=0.9\linewidth]{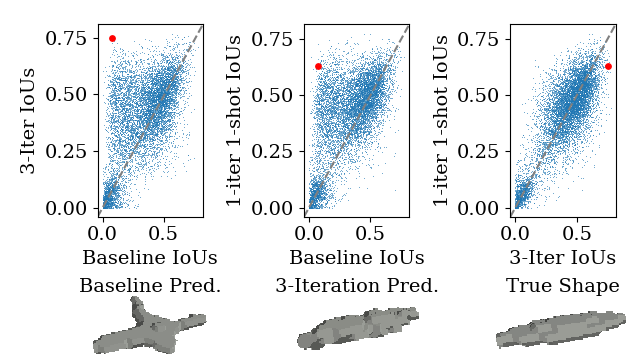}
\end{center}
   \caption{Scatter plots of model performances vs each other for \textbf{vessels}. Note that a point on the identity line has equal reconstruction IoU across two models. Predictions from the Baseline and 3-Iteration models for the red datapoint are shown in the bottom row. A 10-shot prior was used.}
\label{fig:iou_scatters}
\end{figure}

% The above analysis describes the \textit{distribution} of the performances.
Having analyzed the distribution of performances, we now graph the relations between model performances on the same input in Figure~\ref{fig:iou_scatters}.
% We now look at the relations between model performances on the same input via scatterplots in Figure~\ref{fig:iou_scatters}.
We see that our models improve upon baseline performance for the vast majority of datapoints. 
We confirm that our new models mitigate many bad predictions, evidenced by clusters where the Baseline IoU is approximately 0.2 while our models achieve double that or more. 

Figure~\ref{fig:iou_scatters} also shows an example instance for which the reconstruction changes significantly, and demonstrates the cause of this performance difference.
 \textbf{Vessels} are very elongated, and the only elongated category in the training set is  \textbf{airplanes}. 
However,  \textbf{airplanes} have wings and  \textbf{vessels} do not.
The baseline, relying on the prior it has learnt on  \textbf{airplanes}, erroneously includes the wings in the reconstruction.
In contrast, our model uses the provided prior to avoid this mistake.
It is important to note that our model does this \emph{without any retraining}, simply by virtue of disentangling the prior from other aspects of the reconstruction problem.

The previous discussion also suggests that improvements on different categories should vary depending on how far the class is from the set of base categories in terms of its distribution of shapes.
To see if this is true, we present the per-category accuracies of the baseline, 3-iteration full-prior, and 1-iteration 1-shot models in Table~\ref{table:category_performances}.
We see that both of the new models perform impressively on \textbf{rifles} and \textbf{vessels} and neutrally to poorly on \textbf{cabinets}.
Referring back to the average shapes presented in Figure~\ref{fig:average_voxels}, we note that  \textbf{vessels} and \textbf{rifles}, the two categories that our models perform best on, are both very elongated.
The only elongated category in the training set is  \textbf{airplanes}.
Meanwhile,  \textbf{cabinets} have a simple blocky prior. We hypothesize that this makes the prior less useful for learning, as such a basic shape is very simple to extrapolate from an image.% Indeed, we see that our model offers the highest performance increases on categories that the baseline does poorly at.

\begin{table}[b]
\begin{center}
\resizebox{0.99\linewidth}{!}{
\begin{tabular}{ c c c c c }
\toprule
Category & Baseline & 3-Iteration& {1-Iteration 1-Shot} &  {1-Shot Guess} \\
\midrule
Benches & 0.37 & 0.39 (5.4\%) &  0.37 (0.0\%) & 0.13 (-64\%) \\
Cabinets & 0.66 & 0.62 (-6.5\%) &  0.66 (0.0\%) & 0.29 (-56\%)\\
Lamps & 0.18 & 0.19 (5.6\%)  & 0.19 (5.6\%) & 0.11 (-40\%) \\
Sofas & 0.50 & 0.51 (2.0\%) & 0.52 (4.0\%) &  0.33 (-35\%) \\
Vessels & 0.33 & 0.37 (12\%) &  0.38 (15\%) & 0.22 (-34\%)\\
Rifles & 0.12 & 0.16 (33\%)  & 0.19 (58\%) & 0.27 (120\%) \\
\midrule
Mean & 0.36 & 0.37 & 0.39 & 0.23\\
\bottomrule
\end{tabular}%
}
\end{center}
   \caption{Per-category transfer performances. A 1-shot prior was used for both models. The far-right column is the result of naively guessing a random shape from the training set. 
The accuracy of our models are correlated with the accuracy of the 1-shot guess, yet avoid large errors when 1-shot guesses are very poor.}
   \label{table:category_performances}
\end{table}

\subsubsection{Importance of the prior}
A neccessary question to ask when implementing a prior for a model is whether the observed performance stems from model or the prior itself. 
One could hypothesize that the improved results we see could be due to the model simply regurgitating the input prior.
To test this, we performed experiments with a naive baseline that simply outputs the prior without taking into account the image at all.
In the far-right column of Table~\ref{table:category_performances}, we show the average IoU for such a baseline using a 1-shot prior. 
We see that, while the performance of this naive guess is \emph{correlated} with our model in terms of its difference from baseline performance, it performs significantly worse than both of our models.
We also tested the performance of the naive prior guess with up to 25 shot priors, never observing category-wise IoU greater than 0.30.
This shows that our model does provide valuable inference, and it is the combination of the prior with this inference that yields the performance.

At the other extreme of prior quality, we experimented with using the target shape as the input prior, where the 1-iteration 1-shot model achieved an IoU of 0.64 on the training categories and 0.41 on the novel categories 
This might be because the network is combining the provided prior with both the image information as well as general shape priors it has learnt from other classes, this is indeed intended behavior.
Finally, we note that using different 1-shot shapes on the same image-target pair results in a score distribution with $\sigma \approx 0.05$.

\begin{figure}[t]
\begin{center}
\includegraphics[width=0.9\linewidth]{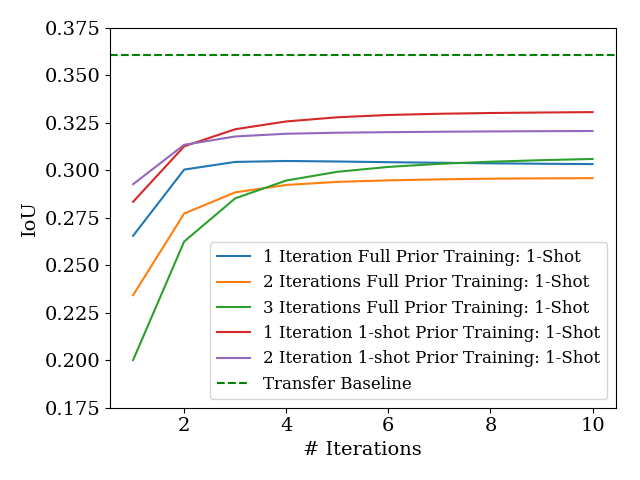}
\end{center}
   \caption{Performance across iterations with the model being fed a 1-shot prior from a random training/testing category.  The green line is the transfer baseline of 0.36. We see that the models never achieve baseline performance, confirming the neccessity of categorical knowledge when implementing the presented framework.
%Interestingly the training-category performances do not decrease as in Figure~\ref{fig:iterative_performance}, instead they increase to a plateau. Observe that the 1-iteration 1-shot model actually achieves the best performance and improves with iteration despite not doing this when given an accurate prior.
}
\label{fig:randomized_input_iterative_performance}
\end{figure}

\paragraph{Performance With Inaccurate Priors}
An assumption of our framework is categorical knowledge at runtime, allowing the selection of a prior shape.
As we have shown, this assumption enables boosted performance on novel categories.
In Figure~\ref{fig:randomized_input_iterative_performance} we perform experiments to observe what happens when that assumption breaks down.
We run our model on the novel categories with priors drawn from other randomly selected categories.

We see that the models never achieve baseline performance, implying that categorical information is necessary to obtain the improvement that we have seen. 
This might be construed as a disadvantage of the presented framework, but it is also evidence that the model has disentangled the category-specific and the category-agnostic aspects of the problem and is relying more on the input prior to provide the category-specific information.
%This suggests that the model is relying on the prior to provide it th
%This is a disadvantage of the presented framework: when a model is fed categorical information during training it may not extract it implicitly itself, which can result in large errors when given incorrect inputs.
In practice, given the advanced state of image classification, knowledge of the category at test time is a valid assumption.
This assumption is in fact common in prior single-view reconstruction work~\cite{yang, atlasnet}.
%such a system would ideally be used in an environment where categories are available at inference time, or while paired with an image classifier. Given the advanced state of image classification, this system could still have real-world applicability.
% In practice, such a system would ideally be used in a controlled environment where categories are always known. Given the advanced state of image classification, even in the few-shot domain, this system could still have practical real-world use when paired with a powerful image classifier. 
% Note that this breaks the paradigm of end-to-end learning that is typical in modern machine learning systems, suggesting that more modular tailored systems can indeed offer increased performance on tasks.

It is interesting to note that the performance of the 1-iteration model trained with 1-shot priors suffers substantially less than other models on the transfer task when given an incorrect prior. We hypothesize that, given the high variability of 1-shot input priors, this model has come to rely less on the prior than others. 
%It is interesting to note that whle the 1-iteration model trained with 1-shot priors does not achieve baseline peformance either, it performs substantially better than other models on the transfer task when given an incorrect prior.
%We hypothesize that this robustness to bad priors is due to the high-variability of 1-shot input priors in training; the model has been encouraged to not fully rely upon a prior. 

\subsubsection{Application to In-the-Wild Images}

% We will include these in the supplementary material upon acceptance.
% We use a model pre-trained on ShapeNet and subsequently finetune it on PASCAL 3D+.
We finetune the 1-shot 1-iteration model on PASCAL 3D+ \cite{xiang2014beyond}.
We train on all 13 ShapeNet categories and 7 of the 10 non-deformable PASCAL 3D+ categories.
We hold out \textbf{bicycles, motorcycles} and \textbf{trains} as these categories are not present in the ShapeNet dataset. 
As seen in Table\ref{table:pascal} our model far outperforms the image-only architecture on both the training and testing categories. 
These results should be considered cautiously due to extremely low variation of PASCAL models, as noted in the original PASCAL 3D+ paper \cite{xiang2014beyond}.
As observed by Tatarchenko et al., retrieval techniques work extremely well on PASCAL, explaining why a shape prior is so useful\cite{tatarchenko2019single}.

% We attribute this difference to the value of the shape prior increasing for the challenging in-the-wild images as well as the low variation of PASCAL models.
% It should be noted in interpreting these results that PASCAL 3D+ has relatively low model diversity and thus is very well-suited to our approach.

\begin{table}
\begin{center}
\resizebox{0.99\linewidth}{!}{
\begin{tabular}{ c c c }
\toprule
Model & Training Categories (Validation) & Novel Categories \\
\midrule
Image-Only & 0.40 & 0.26  \\
\midrule
1-Shot 1-Iteration & 0.50 & 0.37\\
% \midrule
% R2N2-LSTM \cite{r2n2} & 0.44 &0.44 (fully-supervised)\\
\bottomrule
\end{tabular}
}
\end{center}
   \caption{Results of finetuning a ShapeNet-trained model on the common categories of PASCAL3D+. }
   \label{table:pascal}
\end{table}

\section{Future Work}

%This work bridges two fields: 3D reconstruction and few-shot/meta-learning/generalization. We discuss both of these topics and possible further crossover between them.

%\subsection{3D Reconstruction}

The proposed idea of separating out the category-specific prior as an additional input can apply to other single-view reconstruction approaches using other representations of shape too.
%While the task of single-image reconstruction is typically viewed as single-input, we have shown that knowledge of the category can be leveraged. 
%Such knowledge could be explicitly given as we did here, or inferred via a classifier running in parallel to the model.
The prior can be derived from other sources, such as CAD models or geometry-based reasoning.
The results of Tatarchenko et al. also suggest that a simple category-based approach can yield state-of-the-art results on reconstruction, implying a possible crossover between our technique and theirs.
This viewpoint of separating out category-specific priors and learnt category-agnostic refinement can also be applied to many computer vision regression problems (e.g. segmentation or shape completion) that have had relatively little few-shot transfer work done on them.
% In general, our results encourage systems that are not as end-to-end as the standard models in the literature. 

% While we touched on it only briefly, the multiview results here are quite good. The an obvious way forward would be customization of our framework to train in the multiview setting. %Such a system could also be applied to reconstruction from video. 

% The priors that we used here achieved impressive results despite being extremely simple. One could easily imagine other  informative priors, such as geometry-based image reconstructions or bootstrapping from an image-only baseline. % Additionally other inputs could be used in addition to the image-prior pairs.

%\subsection{Few-shot/meta-learning \& Generalization}

% The bulk of few-shot learning research has been on the problem of classification. Computer vision regression problems (e.g. segmentation, depth-estimation, in-painting) have had relatively little few-shot transfer work done on them. These are typically trained end-to-end and could use a similar prior scheme to do few-shot transfer to new-categories. 

% The method we use in the 1-iteration 1-shot method, is similar to that of meta-learning classification approaches that train on the same task they are built for (namely learning from a few examples). These results could be extended to the tasks previously listed as well.

\section{Conclusion}

In conclusion, we presented a new framework for 3D reconstruction that significantly improves generalization to new classes with limited training data, and offers multi-view reconstruction for free. Our models take two inputs: the typical image of the object to reconstruct along with a shape prior. Few-shot knowledge consisting of shape models can be used by inputting the average in as the prior. Such a model can then make iterative predictions by using its own output as a prior. Our model requires no novel class images and no retraining. We identified that our model offers far less extremely poor reconstructions than the baseline. We found that this framework peformed well on the multi-view reconstruction task. This finding in particular is surprising given that this model is never trained on multiview. The results here show that explicit categorical information and priors can be a powerful tool in 3D reconstruction.

{\small
\bibliographystyle{ieee_fullname}
\bibliography{draft}
}

\end{document}